# Sparse Subspace Clustering Friendly Deep Dictionary Learning for Hyperspectral Image Classification

Anurag Goel and Angshul Majumdar, *Senior Member, IEEE*

*Abstract*— Subspace clustering techniques have shown promise in hyperspectral image segmentation. The fundamental assumption in subspace clustering is that the samples belonging to different clusters/segments lie in separable subspaces. What if this condition does not hold? We surmise that even if the condition does not hold in the original space, the data may be nonlinearly transformed to a space where it will be separable into subspaces. In this work, we propose a transformation based on the tenets of deep dictionary learning (DDL). In particular, we incorporate the sparse subspace clustering (SSC) loss in the DDL formulation. Here DDL nonlinearly transforms the data such that the transformed representation (of the data) is separable into subspaces. We show that the proposed formulation improves over the state-of-the-art deep learning techniques in hyperspectral image clustering.

## I. Introduction

ONCE a hyperspectral image is acquired, the task is to label each pixel. However, before such an inference is carried out, the image typically undergoes some preprocessing like denoising, band selection, and so on. There are two approaches for labeling pixels – classification and clustering. Pedagogically there are more studies on classification compared to clustering; even though the latter is more pragmatic. This is because, in the classification-based approach, experts need to manually label a subset of the pixel's values. The labeled pixels are used to train a classifier; once trained, the classifier is used to label the remaining (unlabeled) pixels. The clustering-based approach does not need any manual labeling and is a completely automatic process.

In this work, we address the aforesaid problem, that is, given the image and the number of clusters, we want to automatically label every pixel of the image. Note that clustering is also used in hyperspectral image analysis for band selection [1], [2]; we do not intend to achieve that.

One can find many letters on deep learning-based classification of hyperspectral images but there are only a handful of letters on deep clustering techniques [3]–[8]. All of these studies use an autoencoder (AE) framework for clustering. Letters [3]–[6] use stacked AEs (SAEs) where the representation from the bottleneck layer is input to subspace clustering. In [7], a convolutional AE is used where the representation from the deepest layer is input to $K$-means clustering with Student's $t$-distribution as the distance metric. All of these studies [3]–[7] have the clustering loss in-built into the network. Another work [8] proposes a recurrent neural network-based AE for feature extraction; this study does not incorporate any clustering loss, rather it is an unsupervised representation learning network. The learned representation is input separately to a separate clustering algorithm. These are the only studies on the topic of deep learning-based hyperspectral clustering that we were able to garner from journals published in the last few years. In recent times, variants of classical clustering approaches have also been proposed for the said task. Given the limited scope of this letter, we are unable to discuss them.

One can note that almost all the prior studies are based on the AE/SAE framework. This is because AE and SAE are unsupervised and are amenable to mathematical manipulations which make incorporating the clustering loss a straightforward task. The resulting cost functions are easily solved using gradient descent/backpropagation.

Just like AE/SAE, restricted Boltzmann machine (RBM) or its deep version, the deep Boltzmann machine (DBM) is unsupervised. Theoretically, RBMs and DBMs are more optimal than AEs and SAEs since the former needs to learn only half the number of parameters compared to the latter. Unfortunately, in practice, RBM and DBM cost functions are cumbersome to solve; its training via contrastive divergence does not have the same flexibility as that of backpropagation.

The AE/SAE framework although mathematically flexible is prone to overfitting since one needs to learn the decoder portion along with the encoder. Arguably one can use tied weights, but it has been empirically seen that tied encoder–decoder weights yield poorer results compared to independent encoder and decoder weights. This is the reason that the AE framework tends to overfit in limited data scenarios. In hyperspectral images, if one takes a typical case, e.g., Indian Pines, the number of data points is $145 \times 145$ – of the order of $10^6$. For deep learning, this is not a very large dataset.

Ideally one would prefer a framework that is mathematically flexible and does not need to learn twice the number of parameters. Deep dictionary learning (DDL) [9] satisfies these. DDL has been successfully used in data constrained scenarios for supervised learning [10]–[14]; it has been successfully used for hyperspectral image classification [13], [14] where the number of labeled samples is very small. DDL has not been used for clustering before. This would be the first work that incorporates a clustering loss into the DDL framework. Given the success of subspace clustering in the SAE-based deep learning framework [3]–[6], we will incorporate sparse subspace clustering (SSC) [15] into DDL.

SSC assumes that the samples belonging to different clusters are separable into subspaces. This may not always hold on to the raw data. We postulate that by learning a representation (of the data) via DDL, it may be possible to separate the learned representation into subspaces.

## II. PROPOSED APPROACH

In a regular feedforward deep neural network (assuming two-layers), the data $X$ is projected by the network weights $W_1$ and $W_2$ to a representation $Z$ via some nonlinear activation $\varphi$; the relationship between the output and the representation is

$$\varphi(W_2\varphi(W_1X)) = Z. \quad (1)$$

One can see that solving (1) via backpropagation leads to the trivial solution $W_1 = 0$, $W_2 = 0$, and $Z = 0$.

The situation does not change when an unsupervised clustering loss is incorporated into the solution. For example, let us consider the SSC formulation [15]. In this, it is assumed that the representations belonging to the same cluster lie in the same subspace. The formulation for SSC is as follows:

$$\sum_i \|z_i - Z_{i^c}c_i\|_2^2 + \lambda\|c_i\|_1 \quad \forall i \text{ in } \{1, \ldots, m\}. \quad (2)$$

Assuming $m$ samples, $z_i$ is the representation for the $i$th sample, $Z_{i^c}$ represents all the other representations barring the $i$th one, and $c_i(\in \mathbb{R}^{m-1})$ corresponds to the sparse codes that represent coefficients in $Z_{i^c}$ belonging to the same cluster as $z_i$. The $l_1$-norm imposes sparsity on the codes. The sparsity is important because it ensures that the $i$th sample is only represented by the samples of the same cluster; without the $l_1$-norm the $c_i$ is dense and would imply that the $i$th sample is represented by all the remaining samples; this would not allow for further clustering.

Incorporating the SSC cost into a two-layer neural network will lead to

$$\min_{W_1,W_2,Z,(c_i)_i} \underbrace{\|Z - \varphi(W_2\varphi(W_1X))\|_F^2}_{\text{Neural Network}}$$
$$+ \mu\sum_i \|z_i - Z_{i^c}c_i\|_2^2 + \lambda\|c_i\|_1 \quad \forall i \text{ in } \{1,\ldots,m\}. \quad (3)$$

The trivial solution of (3) would be $W_1 = 0$, $W_2 = 0$, $Z = 0$, and $c_i = 0$, $\forall i$. The symbols have already been defined.

To avoid the trivial solution, prior studies preferred incorporating clustering losses into the AE framework. When one incorporates a clustering loss into an SAE the cost function turns out to be

$$\min_{W_1,W_2,W_1',W_2',(c_i)_i} \underbrace{\|X - W_1'\varphi(W_2'\varphi(W_2\varphi(W_1X)))\|_F^2}_{\text{Stacked Autoencoder}}$$
$$+ \mu\underbrace{\sum_i \|z_i - Z_{i^c}c_i\|_2^2 + \lambda\|c_i\|_1}_{\text{SSC}} \quad \forall i \text{ in } \{1,\ldots,m\}. \quad (4)$$

Here $W_1'$ and $W_2'$ are decoders corresponding to the encoders $W_1$ and $W_2$. One can see that the SSC incorporated AE (4) does not end up in a trivial solution. This is the reason all prior studies [3]–[8] based their formulations on the SAE architecture.

Note that in a regular neural network one needs to learn half the number of weights (only encoder) compared to AEs (both encoder and decoder). The necessity to learn more weights may lead to overfitting which in turn can hamper the generalization ability of the solution. As mentioned in the introduction, one could ideally incorporate clustering loss into the RBM or DBM, but there would be issues in minimizing the resulting cost function.

Owing to the limitations of traditional deep learning architectures, we propose to formulate our solution on DDL. The basic model for two-layer DDL is

$$X = D_1\varphi(D_2Z). \quad (5)$$

Here $D_1$ and $D_2$ represent two layers of dictionaries, $X$ is the data, and $Z$ is the corresponding representation. Note that in DDL, there is no activation in the final layer. This is because the data are real-valued and hence, squashing the output of the network would not map to the real-valued data.

DDL is mathematically flexible which allowed us to integrate SSC into it. This leads to our proposed formulation

$$\min_{D_1,D_2,D_3,Z,H} \underbrace{\|X - D_1D_2D_3Z\|_F^2}_{\text{DDL}}$$
$$+ \mu\underbrace{\sum_i \|z_i - Z_{i^c}c_i\|_2^2 + \lambda\|c_i\|_1}_{\text{SSC}} \quad \forall i \text{ in } \{1,\ldots,m\}$$
$$\text{s.t. } \underbrace{D_2D_3Z \geq 0, D_3Z \geq 0, Z \geq 0}_{\text{ReLU activation}}. \quad (6)$$

Note that we have used rectified linear unit (ReLU) activation here. This is largely owing to two reasons. First, for its ease of solution. Second, for its function approximation ability [16], [17].

We solve (6) using alternating minimization. This leads to the following subproblems:

$$D_1 \leftarrow \min_{D_1}\|X - D_1D_2D_3Z\|_F^2$$
$$D_1^k = XZ_1^\dagger, \text{where } Z_1 = D_2^{k-1}D_3Z^{k-1} \quad (7)$$
$$D_2 \leftarrow \min_{D_2}\|X - D_1D_2D_3Z\|_F^2$$
$$D_2^k = (D_1^k)^\dagger XZ_2, \quad \text{where } Z_2 = D_3^{k-1}Z^{k-1} \quad (8)$$
$$D_3 \leftarrow \min_{D_3}\|X - D_1D_2D_3Z\|_F^2$$
$$D_3^k = (D_1^kD_2^k)^\dagger X(Z^{k-1})^\dagger \quad (9)$$
$$Z \leftarrow \min_Z\|X - D_1D_2D_3Z\|_F^2 + \mu\|Z - ZC\|_F^2 \quad (10)$$
$$c_i^k \leftarrow \min_{c_i}\|z_i - Z_{i^c}c_i\|_2^2 + \lambda\|c_i\|_1 \quad \forall i. \quad (11)$$

**Algorithm 1** DDL + SSC
---
Initialize: $D_1^0, D_2^0, D_3^0, Z^0, C^0$
Repeat till convergence
$D_1^k = XZ_1^\dagger$, where $Z_1 = D_2^{k-1} D_3 Z^{k-1}$
$D_2^k = (D_1^k)^\dagger X Z_2$, where $Z_2 = D_3^{k-1} Z^{k-1}$
$D_3^k = (D_1^k D_2^k)^\dagger X (Z^{k-1})^\dagger$
Update $Z^k$ by solving Slyvester equation -
$(D_1 D_2 D_3)^T (D_1 D_2 D_3) Z + Z(\mu I - \mu C) = (D_1 D_2 D_3)^T X$
Solve $c_i^k$ using SPGL1: $\min_{c_i} \|z_i - Z_{i^c} c_i\|_2^2 + \lambda \|c_i\|_1, \forall i$
End
Compute affinity matrix: $A = |C| + |C|^T$
Use N-cuts to segment $A$

Here "$k$" represents the iteration number. One can see that the dictionary updates are straightforward pseudoinverses. In the update for the sparse codes $c_i$'s, we use the spectral projected gradient solver.[1] For updating $Z$ we take the gradient of (10) and equate it to 0

$$\nabla(\|X - D_1 D_2 D_3 Z\|_F^2 + \mu \|Z - ZC\|_F^2) = 0$$
$$\Rightarrow (D_1 D_2 D_3)^T (D_1 D_2 D_3) Z + Z(\mu I - \mu C)$$
$$\quad - (D_1 D_2 D_3)^T X = 0$$
$$\Rightarrow (D_1 D_2 D_3)^T (D_1 D_2 D_3) Z + Z(\mu I - \mu C) = (D_1 D_2 D_3)^T X.$$

The solution to $Z$ turns out to be via Sylvester's equation of the form $AW + WB = E$ where $W = Z$, $(D_1 D_2 D_3)^T X = C$, $(D_1 D_2 D_3)^T (D_1 D_2 D_3) = A$ and $(\mu I - \mu C) = E$.

The algorithm proceeds by iterative solving for the dictionaries using (9)–(11), updating the coefficients by solving Sylvester's equation and updating the sparse codes $c_i$'s by SPGL1. The problem (8) is nonconvex. Hence, one can at best expect to reach a local minimum; however, we do not have any theoretical guarantees regarding convergence. In practice, we stop the iterations when the values of $c_i$'s do not change significantly with iterations. We emphasize on $c_i$'s since it has a direct consequence on the clustering performance.

Assuming that there are $m$ pixels in the hyperspectral image, $c_i \in \mathbb{R}^{m-1 \times 1}$; this is because the $i$th patch has been omitted while estimating $c_i$. For the sake of uniformity, the corresponding position in $c_i$'s must be imputed with 0; the thus obtained code is $\tilde{c}_i \in \mathbb{R}^{m \times 1}$. These codes $\tilde{c}_i$s are stacked as columns of a matrix $C \in \mathbb{R}^{m \times m}$. The affinity matrix is computed from $C$ using

$$A = |C| + |C|^T. \tag{12}$$

The affinity matrix is segmented using normalized cuts for obtaining the clusters. The complete algorithm is shown in a succinct fashion below.

## III. Experimental Evaluation

### A. Dataset and Experimental Setup

We evaluate our proposed technique on two benchmark datasets – Indian Pines[2] and Pavia University.[3] The standard preprocessing steps are performed on these datasets before classification.

1) The Indian Pines dataset was collected by the Airborne Visible/Infrared Imaging Spectrometer in Northwestern Indiana, with a size of 145 × 145 pixels with a spatial resolution of 20 m per pixel and 10-nm spectral resolution over the range of 400–2500 nm. As is the usual protocol, the work uses 200 bands, after removing 20 bands affected by atmospheric absorption. There are 16 classes.
2) This Pavia University dataset is acquired by a reflective optics system imaging spectrometer (ROSIS). The image is of 610 × 340 pixels covering the Engineering School at the University of Pavia, which was collected under the HySens project managed by the German Aerospace Agency (DLR). The ROSIS-03 sensor comprises 115 spectral channels ranging from 430 to 860 nm. In this dataset, 12 noisy channels have been removed and the remaining 103 spectral channels are investigated in this letter. The spatial resolution is 1.3 m per pixel. The available training samples of this dataset cover nine classes of interest.

We benchmark with four state-of-the-art deep clustering techniques – Deep Spatial-Spectral Subspace Clustering (DS$^3$C) [3], Deep Clustering With Intraclass Distance (DCID) [4], Self-Supervised Deep Subspace Clustering (S$^2$DSC) [5], and 3-D Convolutional AEs (3DCAEs) [7]. These studies have compared with a plethora of shallow and deep clustering techniques and have shown to improve over them. Thus, we only compare with these state-of-the-art clustering methods.

Our proposed technique requires the specification of two parameters $\mu$ and $\lambda$. The parameter $\mu$ controls the relative importance of the dictionary learning and clustering terms. Since there is no reason to favor one over the other, we keep $\mu = 1$. The parameter $\lambda$ determines the sparsity level of the linear weights; we keep $\lambda = 1$ throughout. Our algorithm requires the specification of the number of dictionary atoms. Starting with the input dimensionality dictated by the size of the features, we reduce the number of atoms by half in every stage.

In prior studies on DDL-based hyperspectral image analysis [13], [14], it was found that spectral–spatial features generated by taking principal component analysis (PCA) around the pixel of interest turn out to be a good input feature for DDL. Here we do the same. The process is schematically shown in Fig. 1. Around each pixel, a spatial window is considered; within this window, all the bands in the spectral direction are taken. Each such 3-D patch is then vectorized; these are stacked as columns of a matrix. On this matrix, PCA is run to reduce the dimensionality. We have tried three different windows of sizes 3 × 3, 5 × 5, and 7 × 7; for each window size, 10% of the principal components were kept. For example, in Indian Pines with a 3 × 3 window, the size of the input to PCA would be 1800 (3 × 3 × 200); after PCA 180 principal components will be kept.

Two types of metrics are considered. The first two are generic metrics for clustering where the ground truth is available – normalized mutual information (NMI), adjusted rand index (ARI), purity, and entropy. Since the number classes are assumed to be known for each dataset, the purity and entropy are only computed for those. The second three are specific for hyperspectral image classification – overall accuracy (OA), average accuracy (AA), and Kappa coefficient ($K$); although these three are more suited for classification, studies on clustering also compute these [4], [5]. The results are shown in Table I. For our proposed method, the depth is kept fixed at 3. The experiments were carried out on a 64 bit Intel Core i5-8265U CPU at 1.60 GHz, 16 GB RAM running Ubuntu.

---
[1] https://www.cs.ubc.ca/~mpf/spgl1/index.html
[2] https://paperswithcode.com/dataset/indian-pines
[3] https://paperswithcode.com/dataset/pavia-university

TABLE I
COMPARISON WITH STATE-OF-THE-ART TECHNIQUES

| Dataset | Metric | DS$^3$C | DCID | S$^2$DSC | 3DCAE | Proposed 3x3 | Proposed 5x5 | Proposed 7x7 |
|---|---|---|---|---|---|---|---|---|
| Pavia University | NMI | .637 | .664 | .648 | .653 | **.681** | **.673** | .655 |
| | ARI | .501 | .529 | .507 | .514 | **.550** | **.537** | .518 |
| | Purity | .613 | .694 | .645 | .647 | **.703** | **.700** | .649 |
| | Entropy | .457 | .442 | .451 | .447 | **.404** | **.419** | .444 |
| | OA | .858 | .883 | .866 | .875 | **.897** | **.888** | .875 |
| | AA | .820 | .855 | .831 | .839 | **.862** | **.858** | .841 |
| | Kappa | .791 | .808 | .793 | .796 | **.825** | **.814** | .798 |
| Indian Pines | NMI | .604 | .701 | .685 | .631 | **.726** | **.711** | .696 |
| | ARI | .490 | .523 | .508 | .497 | **.535** | **.531** | .524 |
| | Purity | .607 | .658 | .633 | .637 | **.664** | **.660** | .651 |
| | Entropy | .433 | .426 | .436 | .423 | **.415** | **.418** | .430 |
| | OA | .811 | .841 | .835 | .832 | **.852** | **.846** | .837 |
| | AA | .775 | .801 | .796 | .790 | **.823** | **.810** | .792 |
| | Kappa | .749 | .772 | .770 | .762 | **.790** | **.783** | .762 |

TABLE II
ABLATION STUDIES

| | | One layer Joint | One layer Piecemeal | Two layers Joint | Two layers Piecemeal | Three layers Joint | Three layers Piecemeal | Four layers Joint | Four layers Piecemeal |
|---|---|---|---|---|---|---|---|---|---|
| Pavia University | NMI | .633 | .627 | .654 | .636 | **.673** | .649 | .667 | .651 |
| | ARI | .515 | .509 | .524 | .513 | **.537** | .520 | .533 | .523 |
| | Purity | .611 | .605 | .662 | .614 | **.700** | .657 | .688 | .660 |
| | Entropy | .460 | .468 | .431 | .457 | **.419** | .434 | .425 | .432 |
| | OA | .849 | .832 | .863 | .842 | **.888** | .852 | .880 | .853 |
| | AA | .822 | .809 | .841 | .820 | **.858** | .829 | .851 | .832 |
| | Kappa | .787 | .768 | .802 | .775 | **.814** | .784 | .809 | .785 |
| Indian Pines | NMI | .690 | .685 | .699 | .691 | **.711** | .695 | .706 | .697 |
| | ARI | .511 | .498 | .516 | .509 | **.531** | .516 | .528 | .519 |
| | Purity | .636 | .630 | .651 | .636 | **.660** | .650 | .655 | .651 |
| | Entropy | .429 | .434 | .426 | .428 | **.418** | .427 | .421 | .426 |
| | OA | .819 | .810 | .825 | .818 | **.846** | .827 | .836 | .528 |
| | AA | .781 | .775 | .790 | .779 | **.810** | .788 | .805 | .790 |
| | Kappa | .760 | .751 | .768 | .758 | **.783** | .766 | .777 | .767 |

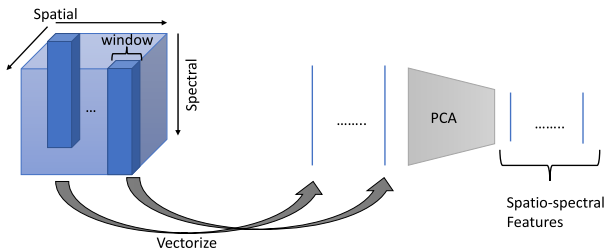

Fig. 1. Spatio-spectral feature extraction.

### B. Results

One can see that our proposed method yields the best results in terms of every possible metric for window sizes $3 \times 3$ and $5 \times 5$. For the larger window size, we perform worse than DCID. The deterioration in results is probably due to overfitting. With a larger window size, the dimensionality increases, and with an increase in dimensionality we need to learn more network weights – this possibly leads to overfitting. Of the existing techniques, DCID yields the best results. In terms of methodology, this is the most sophisticated technique. 3DCAE and S$^2$DSC perform somewhat worse than DCID. DS$^3$C is the simplest approach and consequently yields results that are not at par with the rest.

In the next set of experiments, we carry out ablation studies. We analyze the effect of depth. For each depth, we see how the metrics change for the proposed joint solution and a piecemeal solution. By a piecemeal solution, we mean that the features are generated separately by DDL and the learned features are input to a separate sparse subspace classifier. The joint solution is the proposed one where DDL and clustering losses are intertwined and optimized jointly. The results are shown in Table II for the $5 \times 5$ window.

One observes that the results improve from layers one to three and then dip in layer four for the proposed joint formulation. This is expected; in deep learning, one can improve the results by going deeper; however, one cannot go arbitrarily deep since the number of parameters increases with depth. With limited training data, this leads to overfitting and one sees deterioration in results. For every layer, one can see that the joint formulation yields better results than the piecemeal one. This too is expected; the proposed formulation learns projections that are clustering friendly, but this is not the case for the piecemeal formulation. Overall one can notice that the results from Pavia University are always better than that of Indian Pines. This might be because the former has a considerably higher resolution compared to the latter.

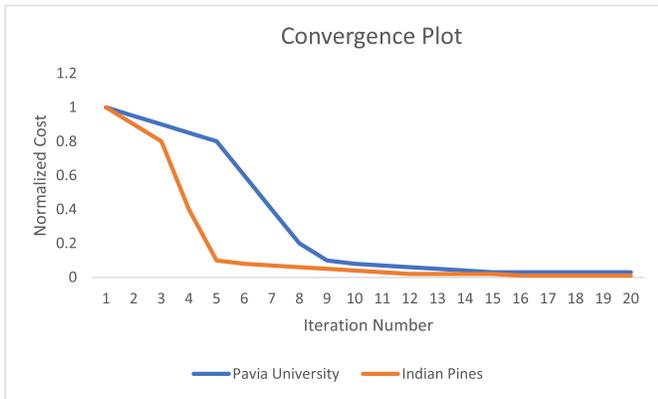

Fig. 2. Empirical convergence plots.

TABLE III
COMPARISON OF RUNTIMES IN SECONDS

| Algorithm | Pavia University | Indian Pines |
|---|---|---|
| DS$^3$C | 1308 | 1221 |
| DCID | 1276 | 1195 |
| S$^2$DSC | 1093 | 1115 |
| 3DCAE | **593** | **320** |
| Proposed (1 layer) | 482 | 303 |
| Proposed (2 layer) | 507 | 346 |
| Proposed (3 layer) | 691 | 406 |
| Proposed (4 layer) | 984 | 793 |

We mentioned before that our formulation is nonconvex and hence, we do not have any convergence guarantees. However, we find that in practice the algorithm converges. The convergence plot for the three-layer $5 \times 5$ window is shown in Fig. 2.

The run times for different algorithms are shown in Table III. The results show that the 3DCAE is the fastest. This is because the said algorithm uses $K$-means clustering internally; $K$-means is faster than subspace clustering which all the other algorithms use. Our proposed technique (at three layers) is slower than 3DCAE but is faster than the rest. Our algorithm for one layer and two layers is faster than 3DCAE but does not yield the best results at these depths.

## IV. CONCLUSION

This work proposes a new approach for hyperspectral image clustering based on the framework of DDL. We incorporate the SSC loss in the DDL framework. Experiments have been carried out on two popular datasets – Pavia University and Indian Pines. We have compared with four state-of-the-art deep learning techniques. Results show that our method is more accurate than others and is also faster than most.

In the future, we would like to incorporate other clustering losses into the DDL framework; namely spectral clustering and $K$-means clustering. We would like to apply these techniques for the problem of hyperspectral band selection via clustering.